%% file: main.tex
\documentclass[10pt]{article} % For LaTeX2e
\usepackage[preprint]{tmlr}
\usepackage{hyperref}
\usepackage{url}

\input{defs.tex}

\title{Effective Neural Network $L_0$ Regularization With BinMask}

\author{\name Kai Jia \email jiakai@mit.edu \\
      \addr Department of Electrical Engineering and Computer Science, \\
      Massachusetts Institute of Technology
      \AND
      \name Martin Rinard \email rinard@csail.mit.edu \\
      \addr Department of Electrical Engineering and Computer Science, \\
      Massachusetts Institute of Technology
      }

\def\month{MM}  % Insert correct month for camera-ready version
\def\year{YYYY} % Insert correct year for camera-ready version
\def\openreview{\url{https://openreview.net/forum?id=XXXX}} % Insert correct link to OpenReview for camera-ready version

\begin{document}

\maketitle
\input{abstract.tex}
\input{allcontent.tex}

\bibliography{refs}
\bibliographystyle{tmlr}

\end{document}

%% file: defs.tex
\usepackage{amsmath}
\usepackage{amssymb}
\usepackage{bm}
\usepackage{mathtools}
\usepackage{algorithm}
\usepackage{algpseudocode}
\usepackage{setspace}
\usepackage{booktabs}
\usepackage{threeparttable}
\usepackage{multirow}
\usepackage{subcaption}
\usepackage[nameinlink,capitalise]{cleveref}
\usepackage[small]{complexity}
\usepackage[inline]{enumitem}

\algrenewcommand\algorithmicrequire{\textbf{Input:}}
\algrenewcommand\algorithmicensure{\textbf{Output:}}
\algdef{SE}[FUNCTION]{Function}{EndFunction}%
[2]{\algorithmicfunction\ \textproc{#1}\ifthenelse{\equal{#2}{}}{}{(#2)}}%
{\algorithmicend\ \algorithmicfunction}%

\newcommand{\real}{\mathbb{R}}
\newcommand{\sciv}[2]{#1\!\times\! 10^{#2}}
\newcommand{\scivv}[1]{10^{#1}}
\DeclarePairedDelimiter{\norm}\lVert\rVert
\DeclarePairedDelimiter{\abs}\lvert\rvert
\DeclarePairedDelimiter{\round}\lfloor\rceil
\DeclarePairedDelimiter{\floor}\lfloor\rfloor
\newcommand{\nth}[1]{#1^{\text{th}}}
\newcommand{\condSet}[2]{\left\{#1 \;\middle|\; #2\right\}}

\newcommand{\V}[1]{\bm{#1}}
\newcommand{\vb}{\V{b}}
\newcommand{\vbsmt}{\hat{\vb}}  % smooth b
\newcommand{\vbr}{\V{b}_r}
\newcommand{\vg}{\V{g}}
\newcommand{\vx}{\V{x}}
\newcommand{\vy}{\V{y}}
\newcommand{\vW}{\V{W}}

\newcommand{\bmlosssym}{\tilde{L}}
\newcommand{\bmloss}[1]{\bmlosssym\left(#1\right)}

\DeclareMathOperator*{\argmin}{arg\,min}
\newcommand{\defeq}{\mathrel{\raisebox{-0.3ex}{$\overset{\text{\tiny def}}{=}$}}}

\newenvironment{enuminline}
{\begin{enumerate*}[label=(\roman*),itemjoin={{; }},itemjoin*={{; and }}]}
{\end{enumerate*}}

%% file: abstract.tex
\begin{abstract}
    $L_0$ regularization of neural networks is a fundamental problem. In
    addition to regularizing models for better generalizability, $L_0$
    regularization also applies to selecting input features and training sparse
    neural networks. There is a large body of research on related topics, some
    with quite complicated methods. In this paper, we show that a
    straightforward formulation, BinMask, which multiplies weights with
    deterministic binary masks and uses the identity straight-through estimator
    for backpropagation, is an effective $L_0$ regularizer. We evaluate BinMask
    on three tasks: feature selection, network sparsification, and model
    regularization. Despite its simplicity, BinMask achieves competitive
    performance on all the benchmarks without task-specific tuning compared to
    methods designed for each task. Our results suggest that decoupling weights
    from mask optimization, which has been widely adopted by previous work, is a
    key component for effective $L_0$ regularization.
\end{abstract}

% vim: spell spelllang=en

%% file: allcontent.tex
\input{introduction.tex}
\input{related.tex}
\input{method.tex}
\input{experiments.tex}
\input{conclusion.tex}

%% file: introduction.tex
\section{Introduction}

Regularization is a long-standing topic in machine learning. While $L_1$ and
$L_2$ regularization have received thorough theoretical treatment and been
widely applied in machine learning, $L_0$ regularization is still under active
investigation due to the optimization challenges caused by its combinatorial
nature. Given a vector $\vW\in\real^n$, $L_0$ regularization penalizes
$\norm{\vW}_0$, the number of non-zero entries in $\vW$. One prominent
application of $L_0$ regularization is to induce sparse neural networks,
potentially improving time and energy costs at inference with minimum impact on
accuracy \citep{chen2020survey}. When applied to the input features, $L_0$
regularization can also select a subset of features important to the learning
task.

Increasingly complicated methods have been proposed for $L_0$ regularization of
neural networks based on various relaxation formulations \citep{
louizos2018learning, yamada2020feature, zhou2021effective}. In this paper, we
revisit a straightforward formulation, BinMask, that multiplies the weights with
deterministic binary masks \citep{jia2020efficient}. The masks are obtained by
quantizing underlying real-valued weights, which are simultaneously updated with
network weights during training. BinMask uses the identity straight-through
estimator to propagate gradient through the quantization operator by treating it
as an identity function during backpropagation. BinMask has demonstrated
empirical success in training low-complexity sparse binarized neural networks
\citep{ jia2020efficient}.

Unlike most research that only focuses on either feature selection or network
sparsification, we evaluate BinMask on three different tasks: feature selection,
network sparsification, and model regularization. We use the same BinMask
implementation without task-specific hyperparameter tuning. BinMask achieves
competitive performance on all three tasks compared to methods designed for each
task. Our results suggest that the decoupled optimization of network weights and
sparsity masks is a key component for effective $L_0$ regularization. When
combined with standard techniques for optimizing binary masks, such a
formulation is sufficient to give impressive results. We will also open source
our implementation, which supports automatically patching most existing PyTorch
networks to use sparse weights with minimum additional code.

% vim: spell spelllang=en

%% file: related.tex
\section{Related work}

\citet{xiao2019autoprune} and \citet{jia2020efficient} are most relevant to our
work. They both use a similar formulation that multiplies weights with
deterministic binary masks optimized via the straight-through estimator. However,
\citet{ xiao2019autoprune} uses bi-level optimization with a custom optimizer to
update weights and masks separately, which is computationally expensive. Their
experiments focus on training sparse networks starting from pretrained networks
without evaluating $L_0$ regularization in other scenarios. \citet{
jia2020efficient} proposes the name BinMask and uses it to train sparse
binarized neural networks with more balanced layer-wise sparsities to achieve
tractable verification. However, they do not evaluate BinMask on other
applications or real-valued networks. By contrast, our method learns weights and
masks simultaneously in each iteration from random initialization, and we
evaluate our method on a diverse set of applications.

Another family of closely related methods exploits probabilistic
reparameterization and relaxation to the $L_0$ ``norm'' \citep{
srinivas2017training, louizos2018learning, zhou2021effective}. They focus on
sparse training and have not evaluated their methods for other applications.
Among them, ProbMask \citep{zhou2021effective} achieves the best accuracies on
sparse training. However, their method is tailored to network sparsification and
is not directly applicable to feature selection or model regularization. A
drawback of methods using probabilistic masks is the need to trade off between
computation time and estimation variance. \citet{louizos2018learning} uses one
sample per minibatch despite larger variance. ProbMask uses multiple samples per
minibatch at the cost of slower training. Sampling of masks also introduces
distribution shift that requires finetuning network weights after fixing masks,
as will be shown in our experiments for ProbMask. By contrast, BinMask uses
deterministic masks, which is faster to train and can deliver well-performing
models without finetuning.

$L_0$ regularization has two critical applications: feature selection and
sparse neural network training. There are enormous methods for both tasks that
do not formulate them as $L_0$ regularization problems. Feature selection
algorithms can be divided into three categories \citep{tang2014feature}. \emph{
Filter models} select features according to ranks computed by some predefined
criterion \citep{gu2012generalized, kira1992practical, vergara2014review}. They
are computationally efficient but deliver worse results due to the ignorance of
the target learner to be used with selected features. \emph{Wrapper models} use
heuristics to select a subset of features that maximize the performance of a
given black-box learner \citep{ el2016review}, which is computationally
expensive because the target learner needs to be evaluated multiple times.
\emph{Embedded models} attempt to address the drawbacks of the other two methods
by simultaneously selecting features and solving the learning task
\citep{liu2005toward, ma2008penalized}. Recently, there has been an increasing
interest in using neural network methods for feature selection
\citep{lu2018deeppink, zhang2019feature, wojtas2020feature, yamada2020feature,
lemhadri2021lassonet}. Some utilize sparse neural networks to derive selected
features from the sparsity patterns \citep{atashgahi2022quick,
atashgahi2023supervised}. \citet{yamada2020feature} is most relevant to our
method, as they multiply the input features of neural networks with stochastic
binary gates. However, they preserve continuous gate variables during training,
which is different from how selected features are used at test time and thus
causes lower performance. Their gate probabilities do not converge in our
experiments.

Sparse neural networks promise faster and more energy-efficient inference,
although there are software/hardware implementation challenges \citep{
lu2019efficient}. Sparse networks can be obtained with diverse pruning methods
that remove neurons from trained dense networks according to different
importance criterion \citep{ mozer1989using, han2015learning, blalock2020state}.
Those methods typically need more computation budget because a dense network
must be trained before pruning. Another line of work learns weights and sparse
connections simultaneously during training, typically via applying learnable
masks on the weights or via thresholding the weights \citep{srinivas2017training,
louizos2018learning, xiao2019autoprune, jia2020efficient, zhou2021effective,
vanderschueren2023straight}. It is also possible to train sparse weights from
random initialization for certain tasks with proper sparse topology \citep{
frankle2018lottery, evci2019difficulty}. Under specific settings, sparse
networks have demonstrated superior performance than dense counterparts
empirically, but the extent and the reason are still under investigation \citep{
jin2022prunings}.

Many related methods, including this one, rely on the Straight-Through Estimator
(STE) to define the gradient for quantization operators that generate the masks.
Despite the lack of rigorous theoretical foundations
\citep{shekhovtsov2022reintroducing}, STE-based methods have demonstrated good
empirical performance in training binarized or quantized neural networks
\citep{hubara2016binarized, krishnamoorthi2018quantizing, bethge2019back,
alizadeh2019empirical, meng2020training}.

% vim: spell spelllang=en

%% file: method.tex
\section{The BinMask method}

\subsection{General formulation}
BinMask applies binary masks to the weights or inputs of neural networks. $L_0$
regularization is accomplished via regularizing the norm of binary masks. This
formulation decouples the optimization of weight values from the optimization of
masks, which allows recovering removed weights to their values before they are
masked out (except changes by weight decay). Such a formulation is favorable
because:
\begin{enuminline}
    \item allowing the growth of connections has been shown to yield better
        sparse networks \citep{evci2020rigging}
    \item recovering weights to previous values is compatible with the finding
        that tuning pruned networks with earlier weight values or learning rate
        schedule is beneficial \citep{renda2020comparing}
\end{enuminline}.

Formally, given a function $f: \real^n \mapsto \real^m$ representing the neural
network with inputs and parameters packed as one argument, a loss functional $L:
(\real^n \mapsto \real^m) \mapsto \real$ (i.e., a higher-order function that
maps another function to a real value, which can include a training procedure of
the neural network), an integer $k \in [1, n]$ denoting the number of values
allowed to be masked out, and a penalty coefficient $\lambda \ge 0$, BinMask
attempts to solve the following optimization problem:

\begin{align}
    \argmin_{\vb\in\{0,\,1\}^k} \; & \bmloss{\vb} \label{eqn:bmloss} \\
    \text{where } & \bmloss{\vb} \defeq L(f \circ g_{\vb}) + \lambda
        \norm{\vb}_1 \nonumber \\
    & g_{\vb}(\vx) \defeq \vx \odot [\vb; \V{1}_{n-k}] \nonumber
\end{align}

Assuming $L(\cdot)$ and $f(\cdot)$ can be evaluated in polynomial time, the
corresponding decision problem of \cref{eqn:bmloss} (i.e., deciding if $\exists
\vb:\; \bmloss{\vb} < c$) is \NPC{} because it takes polynomial time to encode a
3-SAT instance in $L$ or verify a solution. Therefore, exactly solving
\cref{eqn:bmloss} is \NP-hard.

Assuming $\bmloss{\vb}$ is differentiable with respect to $\vb$, we use a
gradient-based optimizer to optimize real-valued latent weights $\vbr \in
\real^k$ that are deterministically quantized to obtain the binary mask $\vb$.
We use the identity straight-through estimator to propagate gradient through the
non-differentiable quantization operator, which has demonstrated good empirical
performance in training binarized or quantized neural networks
\citep{hubara2016binarized, krishnamoorthi2018quantizing}. Formally, we
transform \cref{eqn:bmloss} into the following problem:

\begin{align}
    \argmin_{\vbr \in \real^k} \;& \bmloss{q(\vbr)} \label{eqn:bmloss:real} \\
    \text{where } & q_i(\vb) \defeq \left\{
        \begin{array}{lr}
            1 & \text{ if } \vb_i \geq 0 \\
            0 & \text{ if } \vb_i < 0
        \end{array}
    \right. \nonumber
\end{align}

\cref{eqn:bmloss:real} is solved via a gradient-based optimizer. The identity
straight-through estimator defines the gradient of $q(\cdot)$:
$\frac{\partial \bmlosssym}{\partial \vbr} = \frac{\partial \bmlosssym}{\partial
q(\vbr)}$.

The loss functional $L(\cdot)$ often includes an iterative optimization
procedure of the network weights. We optimize $\vbr$ and neural network weights
simultaneously in such cases. Since it has been suggested that during
initialization, the weights undergo volatile changes \citep{leclerc2020two}, and
pruning at initialization is more challenging~\citep{frankle2021pruning}, we
freeze the binary masks during initial epochs. We initialize the BinMask weights
$\vbr$ with a constant value $\alpha_0$ and clip its values between $[-\alpha_1,
\, \alpha_1]$ during training. To optimize $\vbr$, we use the Adam optimizer
\citep{ kingma2014adam}, whose use of the second moment has been shown to be
crucial for training performant binarized neural networks \citep{
alizadeh2019empirical}. We use cosine learning rate annealing to schedule
learning rates \citep{ loshchilov2017sgdr}. Since $\norm{\vb}_1$ is not
differentiable at $\vb=0$ and $\lambda \norm{\vb}_1 = \frac{1}{2} \lambda
\norm{\vb}_2^2$ for $\vb \in \{0,\, 1\}^k$, we use the quadratic form in the
implementation. \cref{alg:binmask-train} summarizes our algorithm.

\begin{algorithm}[tb]
    \caption{Training with BinMask}
    \label{alg:binmask-train}
\begin{algorithmic}[1]
    \Require Dataset $D$
    \Require Neural network function $f(\vx,\,\vW)$ with
        initial weights $\vW$ and loss function $L(\vy,\,\hat{\vy})$
    \Require Total number of epochs $E$
    \Require Number of iterations per epoch $T$
    \Require An optimizer $U(\vW,\,\nabla)$ for the neural
        network
    \Require A specification $s$ describing to which weights
        and inputs should BinMask be applied
    \Require BinMask regularization coefficient $\lambda$
    \Require BinMask initialization $\alpha_0$ (default:
        0.3)
    \Require BinMask clip $\alpha_1$ (default: 1)
    \Require BinMask initial learning rate $\eta_0$
        (default: $10^{-3}$)
    \Require BinMask final learning rate $\eta_1$
        (default: $10^{-5}$)
    \Require BinMask warmup epoch fraction $E_b$ (default: 0.1)
    \Ensure Trained weights $\vW^*$ and BinMask values $\vb^*
        \in \{0,\,1\}^{|s|}$.

    \setstretch{1.4}

    \State $E_b' \gets \round{E_bE}$
    \State $\vbr \gets \alpha_0\V{1}_{|s|}$
        \Comment{Initialize $\vbr$ to be a constant vector corresponding to
            items described by $s$}
    \For {$e \gets 0$ \textbf{to} $E-1$}
        \If {$e \geq E_b'$}
            \State $\eta_b \gets \eta_0 + \frac{\eta_1-\eta_0}{2} \left(
                \cos\left( \frac{e - E_b'}{E - E_b' - 1} \pi \right) + 1
            \right)$
                \Comment{schedule BinMask learning rate}
        \EndIf

        \For {$t \gets 0$ \textbf{to} $T-1$}
            \State Sample $(\vx,\,\vy) \sim D$
            \State $\vb \gets q(\vbr)$
                \Comment{$q(\cdot)$ defined in \cref{eqn:bmloss:real}}
                \label{alg:binmask-train:before-smooth}
            \State $(\vx',\, \vW') \gets s(\vx,\, \vW,\, \vb)$
                \Comment{Multiply binary masks with the weights or inputs
                    specified by $s$}
            \State $\vg_W \gets \frac{\partial}{\partial \vW}L(f(\vx',\,
                \vW'),\, \vy)$
            \State $\vW \gets U(\vW,\, \vg_W)$
                \Comment{Update the weights with the user-provided optimizer}
            \If {$e \geq E_b'$}
                \State $\vg_b \gets \left(
                    \frac{\partial}{\partial \vb}L(f(\vx',\,
                    \vW'),\, \vy)\right) + \lambda \vb$
                \State $\vbr' \gets \operatorname{Adam}(\vbr,\, \vg_b,\, \eta_b)$
                    \Comment{Get updated BinMask weights with the Adam
                        optimizer}
                \State $\vbr \gets \min\{\max\{\vbr',\, -\alpha_1\},\,
                    \alpha_1\}$
            \EndIf
        \EndFor
    \EndFor
    \State \Return $\vW^* \gets \vW$, $\vb^* \gets q(\vbr)$
\end{algorithmic}
\end{algorithm}

\subsection{Feature selection \label{sec:binmask:fsel}}
We apply BinMask to the inputs of a network for feature selection. Note that if
the first layer is fully connected, applying BinMask to the input is equivalent
to applying BinMask to groups of the first layer weights, which is structural
sparsification of the first layer by requiring that either all or none of the
connections from each input to the neurons must be selected. Unfortunately, the
BinMask algorithm described above is not ideal for practical feature selection
for two reasons:
\begin{enumerate}
    \item The binary mask may change after each iteration, making the result
        sensitive to the number of iterations.
    \item Typical feature selection algorithms allow selecting a given number of
        features requested by the user rather than indirectly controlling the
        size of selected feature set through a regularization coefficient.
        However, it is challenging to control $\lambda$ to select a precise
        number of features with BinMask, especially given the randomness in
        minibatch sampling during training.
\end{enumerate}
To ameliorate those two problems, we introduce a smoothed mask, denoted as
$\vbsmt$, for feature selection with BinMask. We compute the smoothed mask with
moving average after line \ref{alg:binmask-train:before-smooth} in
\cref{alg:binmask-train}:
\begin{align}
    \vbsmt \gets \gamma\vbsmt + (1 - \gamma)\vb
        \label{eqn:smooth-mask}
\end{align}
We initialize $\vbsmt$ with zeros and set $\gamma=0.9$ in our experiments. Let
$\vbsmt(\lambda)$ denote the smoothed mask computed with regularization
coefficient $\lambda$. If the user provides a value of $\lambda$ without
specifying the number of selected features, we select the $\nth{i}$ feature if
$\vbsmt_i(\lambda) \ge 0.5$. Of note, most of the smoothed mask values converge
to either zero or one. When the user requests exactly $k$ features, we need to
find a $\lambda$ value to not arbitrarily cut features in the convergent region.
We search for $\lambda^*$ and a threshold $c_{\lambda^*}$ such that $0.2 \le
c_{\lambda^*} \le 0.8$ and $\abs*{\condSet{i} {\vbsmt_i(\lambda^*) \ge
c_{\lambda^*}}} = k$. We use exponential search starting with $\lambda_0 =
\scivv{-3}$. We select the $\nth{i} $ feature if $\vbsmt_i(\lambda^*) \ge
c_{\lambda^*}$.

% vim: spell spelllang=en

%% file: experiments.tex
\section{Experiments}

We evaluate the effectiveness of BinMask on three tasks: feature selection,
network sparsification, and model regularization. We implement BinMask with a
standalone Python module based on PyTorch \citep{paszke2019pytorch}. Our
implementation can be easily incorporated into other PyTorch projects and
supports automatically patching most existing PyTorch networks to use BinMask
for weight sparsification with minimum additional code. We will open source the
implementation after paper review. We use the same implementation for all the
experiments, with hyperparameters given in \cref{alg:binmask-train} except for
one change in feature selection.

\subsection{Feature selection}
% f{{{

\newcommand{\addSubFig}[2]{
    \begin{subfigure}{.31\textwidth}
        \centering
        \includegraphics[width=\textwidth]{#1}
        \vskip -.5em
        \caption{\footnotesize #2}
    \end{subfigure}
}
\begin{figure}[hbt]
    \centering
    \addSubFig{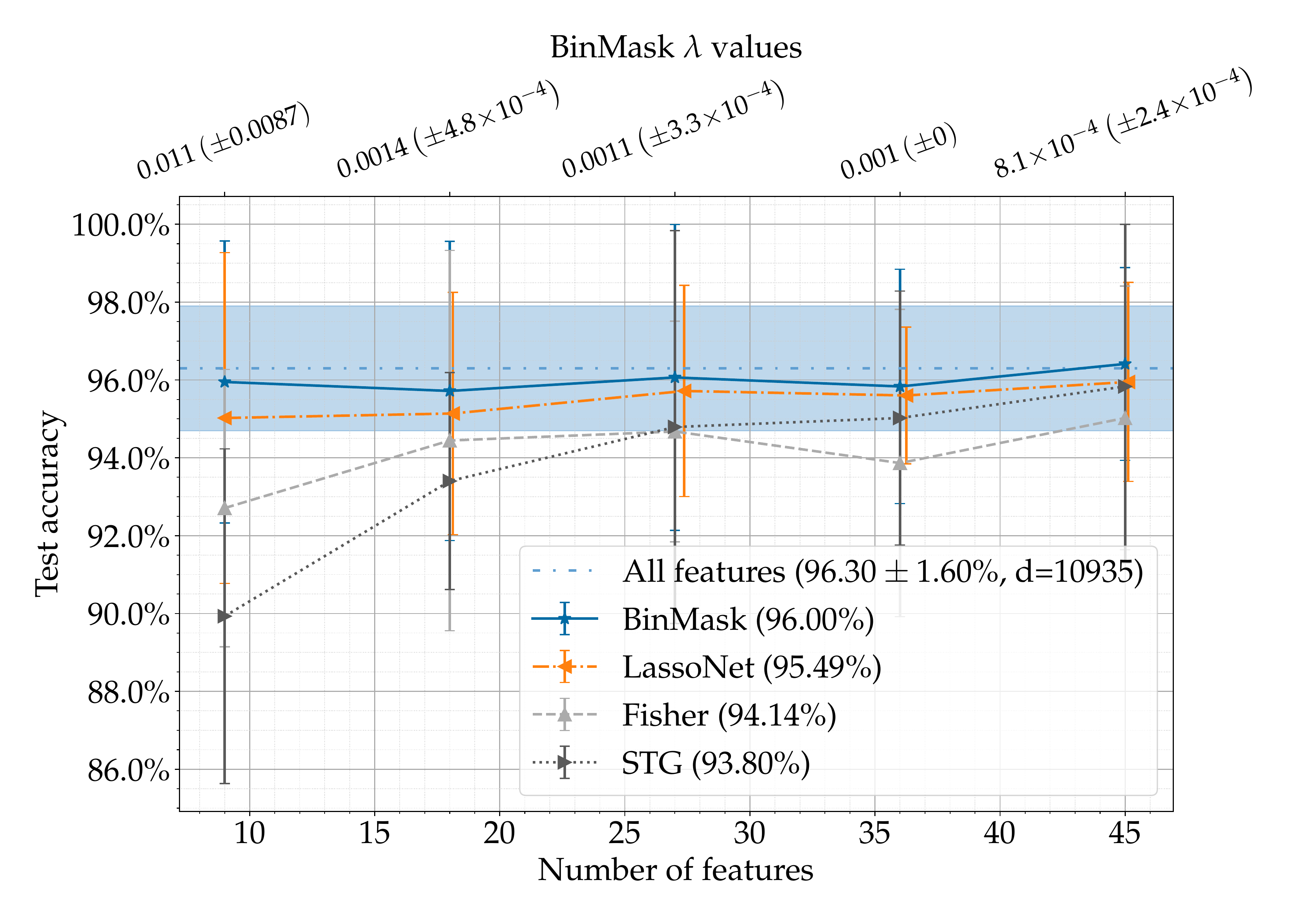}{AP-Breast-Ovary ($n=542$; $c=2$)}
    \hfill
    \addSubFig{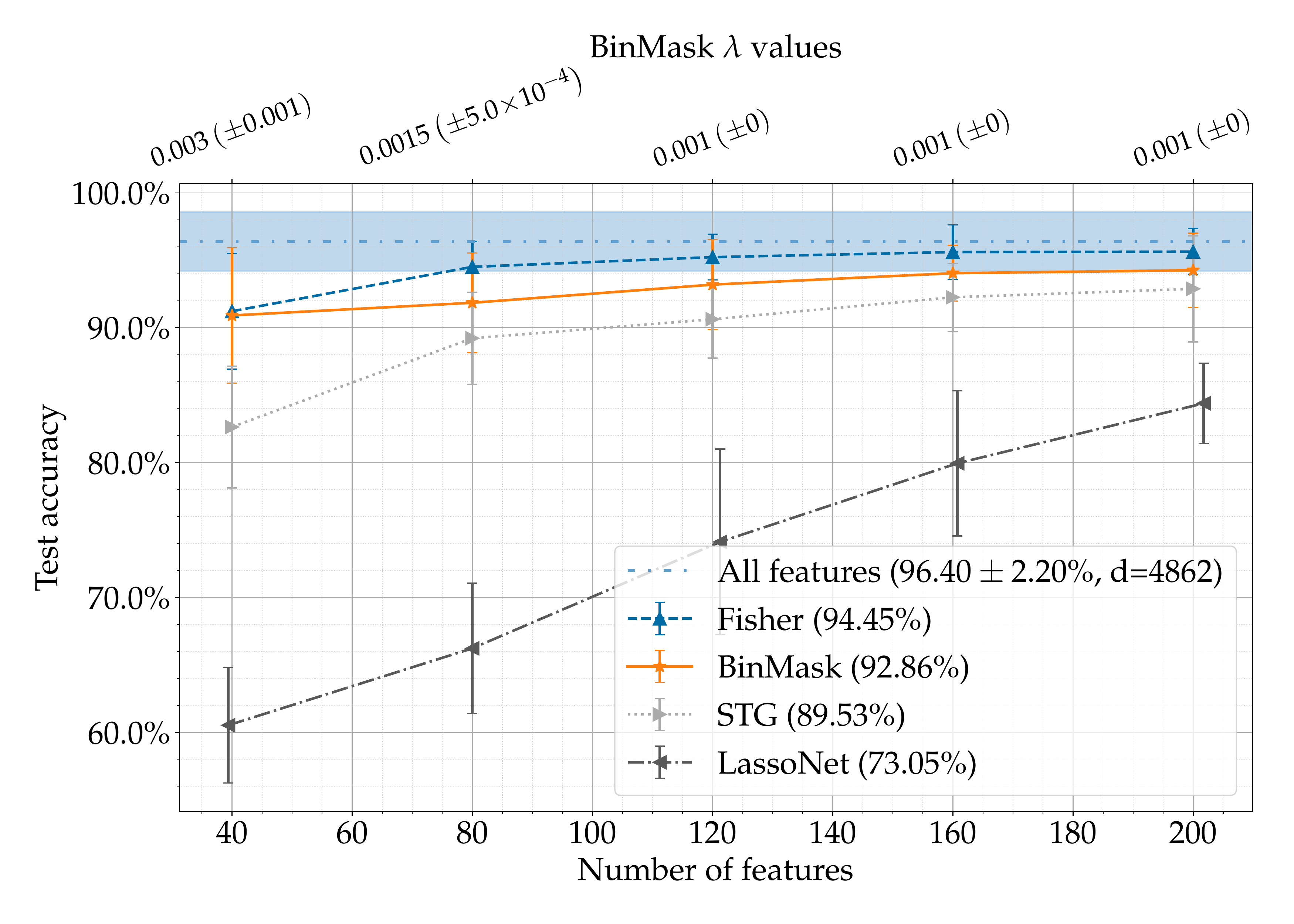}{BASEHOCK ($n=1,993$; $c=2$)}
    \hfill
    \addSubFig{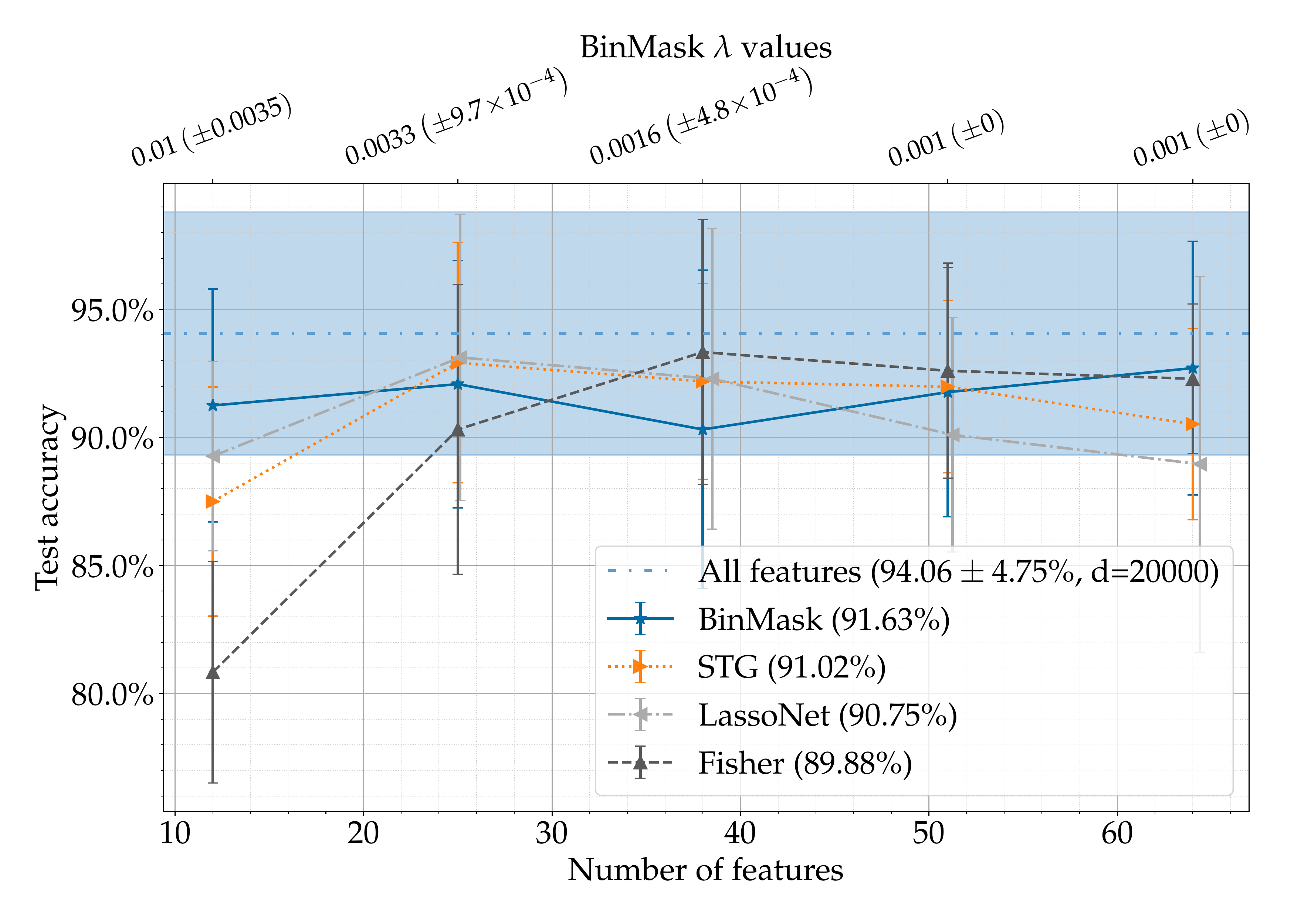}{Dexter ($n=600$; $c=2$)}
    \vskip .8em
    \addSubFig{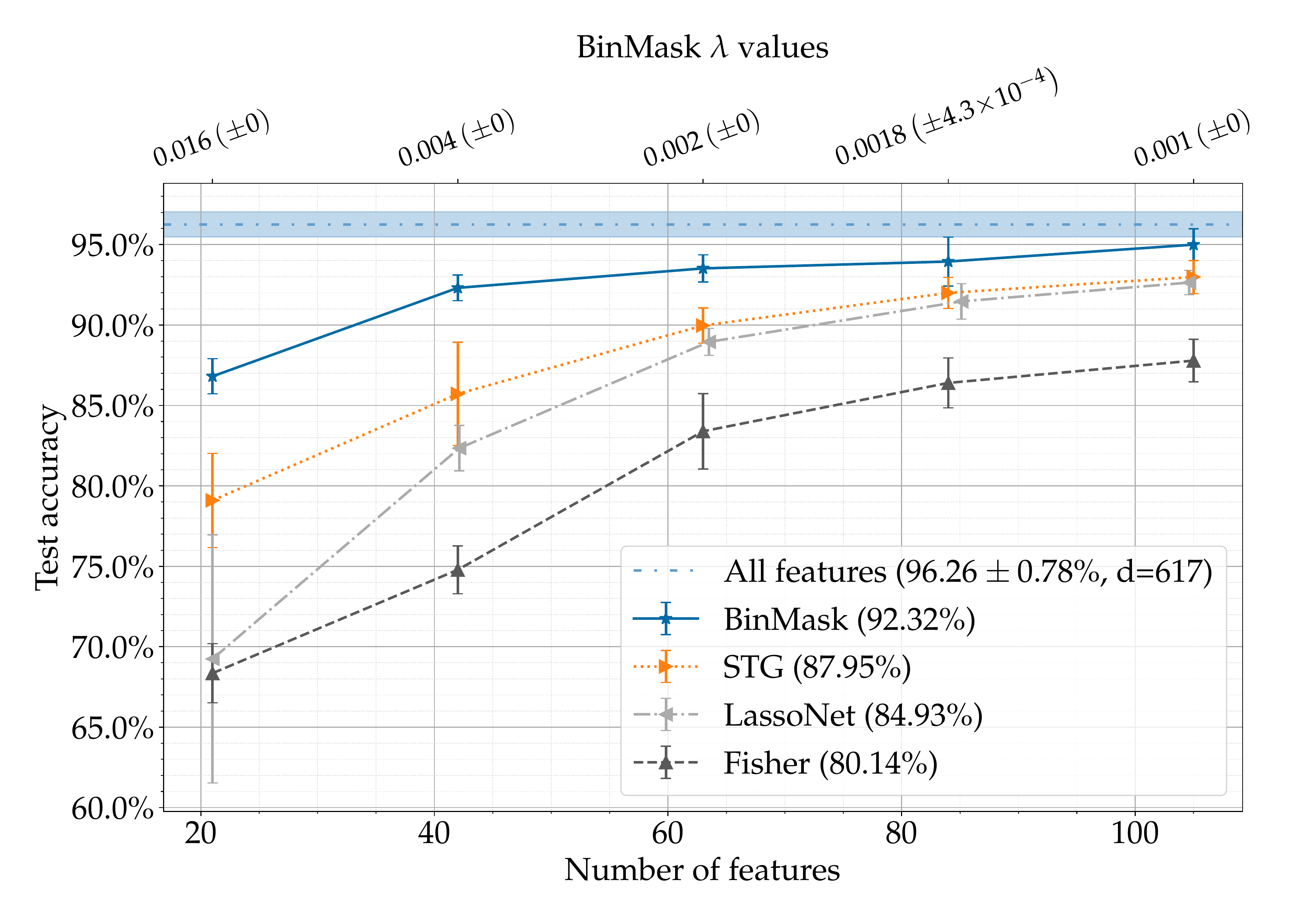}{Isolet ($n=7,797$; $c=26$)}
    \hfill
    \addSubFig{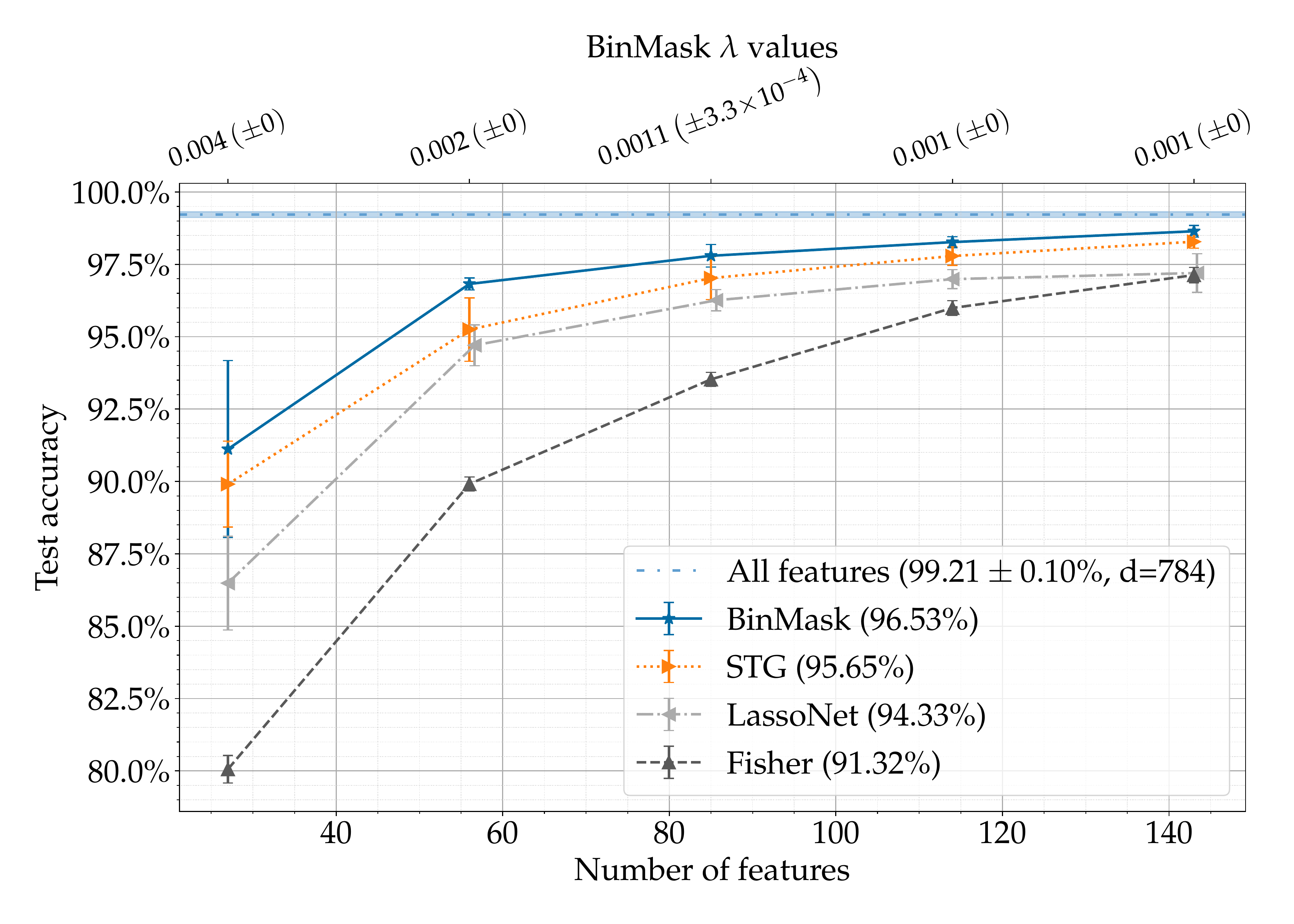}{MNIST-CNN ($n=70,000$; $c=10$)}
    \hfill
    \addSubFig{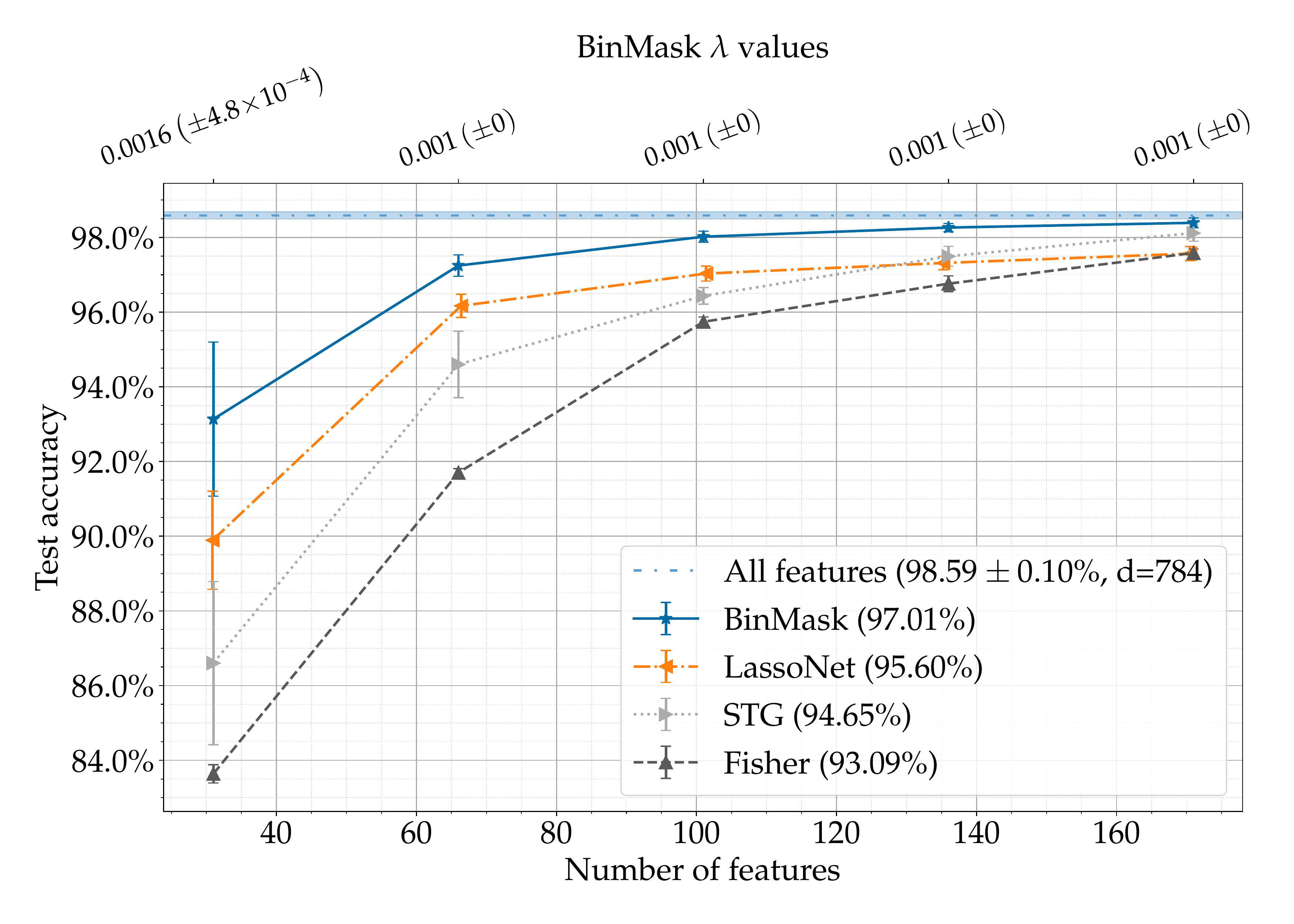}{MNIST-MLP ($n=70,000$; $c=10$)}

    \vskip 1em
    \caption{Feature selection results on various datasets. Bracketed numbers in
        the legends are mean accuracy over all numbers of features. The top
        horizontal axis, BinMask $\lambda$ values, are the values of
        regularization coefficients found by exponential search to select a
        given number of features, where bracketed numbers are standard deviation
        over eight trials. Error bars and the transparent horizontal region
        indicate 95\% CI computed over eight trials. For each dataset, $n$ is
        the total number of data points, $c$ is the number of classes, and $d$
        is the number of input features.}
    \label{fig:fsel}
\end{figure}

\noindent\textbf{Datasets}: We evaluate the capability of BinMask and other
methods to select a subset of features that maximize classification accuracy on
five datasets. AP-Breast-Ovary is a binary classification dataset with gene
expression features. BASEHOCK is a text classification dataset. Dexter is a text
classification task with purposely added noninformative features used as a
feature selection benchmark at NeurIPS 2003 \citep{guyon2004result}. Isolet is a
speech classification dataset. MNIST is a handwritten digit classification
dataset with pixel values as inputs. Note that for MNIST, feature selection is
selecting a set of pixel locations to maintain classification accuracy. Our
comparison methods also use BASEHOCK, Isolet, and MNIST. We select the other
two datasets by querying OpenML \citep{OpenML2013} for classification datasets
with no missing features and at least 512 instances. We then manually check the
top few datasets with the lowest instance-to-feature ratio to select the two
datasets with a nonempty description. Features of each dataset are linearly
normalized to the range $[0,\, 1]$. We use the provided training/test split of
MNIST and randomly select 20\% instances as the test set for other datasets.

\noindent\textbf{Classifiers}: For MNIST, we consider two neural network
architectures. One is an MLP with two hidden layers, each with 512 neurons and
ReLU followed by batch normalization \citep{ioffe2015batch}, denoted as
MNIST-MLP in the results. The other is a LeNet5 convolutional neural network
\citep{lecun1998gradient} with batch normalization after convolutional layers,
denoted as MNIST-CNN. For other datasets, we use an MLP with two hidden layers,
each having 64 and 20 neurons with tanh, respectively. MNIST networks are
trained for 10 epochs and other networks for 100 epochs. For all tasks, we use
the SGD optimizer with momentum 0.9 and weight decay $\sciv{5}{-4}$, with cosine
learning rate annealing from 0.1 to $\scivv{-5}$. Each minibatch has 256
instances. If a dataset is too small, the training set is duplicated to have at
least 30 minibatches per epoch.

\noindent\textbf{BinMask setup}: Since with the default setting of
\cref{alg:binmask-train} and the small training sets, the networks easily
overfit the training data before any feature is masked out, making the gradients
with respect to inputs uninformative for mask optimization, we set the
initialization $\alpha_0=0.02$ to allow easier mask flip before overfitting.
Other hyperparameters use their default values. For each task, we apply BinMask
to the inputs of a network with the same architecture as the classification
network for that task.

\noindent\textbf{Comparison}: We compare BinMask with three other methods.
Fisher score is a classical filter model for supervised feature selection that
aims to find a subset of features to maximize interclass distance while
minimizing intraclass distance \citep{gu2012generalized}. We use the
implementation provided by \texttt{scikit-feature}~\citep{li2018feature}. Fisher
score produces a weight for each feature that can be ranked to select a given
number of features. Stochastic gates (STG) uses a probabilistic relaxation to
regularize the $L_0$ ``norm'' of features while simultaneously learning a
classifier \citep{yamada2020feature}. We use the open-source implementation
provided by the authors\footnote{\url{https://github.com/runopti/stg}}. For each
task, we apply the STG layer to the inputs of a network with the same
architecture as the classification network for that task. We use their default
hyperparameter settings. STG produces probabilities for selecting each input
feature. Since the STG paper does not specify an algorithm to choose a
probability cutoff threshold for feature selection, we use the same exponential
search strategy outlined in \cref{sec:binmask:fsel} to find a suitable
regularization coefficient for STG, starting with the default value $0.1$ given
by their implementation. LassoNet uses a neural network with skip connections
and $L_1$ regularization to simultaneously learn linear and nonlinear components
for the task \citep{lemhadri2021lassonet}. We use the open-source implementation
provided by the authors%
\footnote{\url{https://github.com/lasso-net/lassonet}}. We use their default
hyperparameter settings and a one-hidden-layer neural network, as suggested by
the authors. LassoNet produces a path of networks with decreasing numbers of
features used. We choose from the path the feature set whose size is closest to
the given number of features.

\noindent\textbf{Evaluation}: For a task $t$, we first use BinMask with
$\lambda=0.001$ ($\lambda=\sciv{5}{-4}$ for AP-Breast-Ovary) without feature
number constraint to select a set of features. Let $n_t$ be the number of
features selected. We then run all methods to select $n_t-i\floor{n_t/5}$
features for $0 \le i \le 4$. Given a set of selected features by a method, we
evaluate the accuracy of the classifier network corresponding to the task
trained from new random initialization with selected features. We only keep
selected features as network inputs, except for MNIST-CNN where we use zeros to
replace unselected features so that convolutional layers receive two-dimensional
inputs. We run eight trials with different dataset split and different weight
initialization for each (dataset, method, number of features) configuration. We
evaluate each method by its mean accuracy on the test set over eight trials.
Since BinMask (with smoothed masks) and STG \citep{yamada2020feature} produce
continuously valued masks that should converge to zero or one in the ideal
situation, we also evaluate how likely the computed masks converge. A mask is
defined to have converged if at most 20\% of its values are within $[0.15,\,
0.85]$.

\noindent\textbf{Results}: \cref{fig:fsel} presents the accuracies on the
considered benchmarks, showing that BinMask, a generic $L_0$ regularizer,
exhibits competitive performance compared to other methods tailored to feature
selection. BinMask achieves best average accuracies on five of the six
benchmarks. STG \citep{yamada2020feature} uses 1.6 steps for the exponential
search on average, and 8.86\% of the computed masks converge. By comparison,
BinMask uses 1.7 search steps, with 75.00\% of the masks converging.

% f}}}

\subsection{Network sparsification}
% f{{{

\renewcommand{\addSubFig}[2]{
    \begin{subfigure}{.48\textwidth}
        \centering
        \includegraphics[width=\textwidth]{#1}
        \vskip -.5em
        \caption{#2}
    \end{subfigure}
}
\begin{figure}[hbt]
    \centering
    \addSubFig{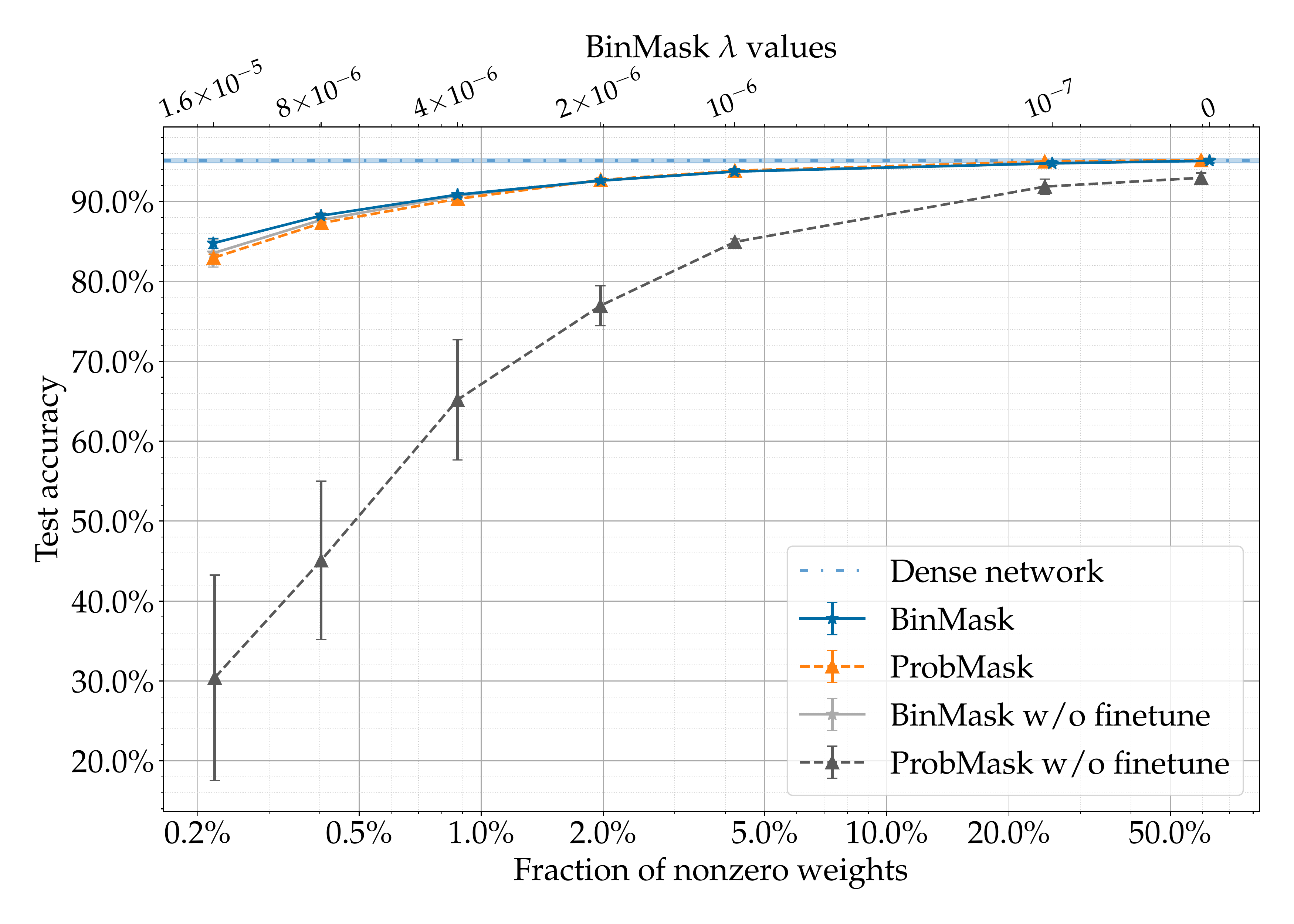}{ResNet32 on CIFAR10}
    \hfill
    \addSubFig{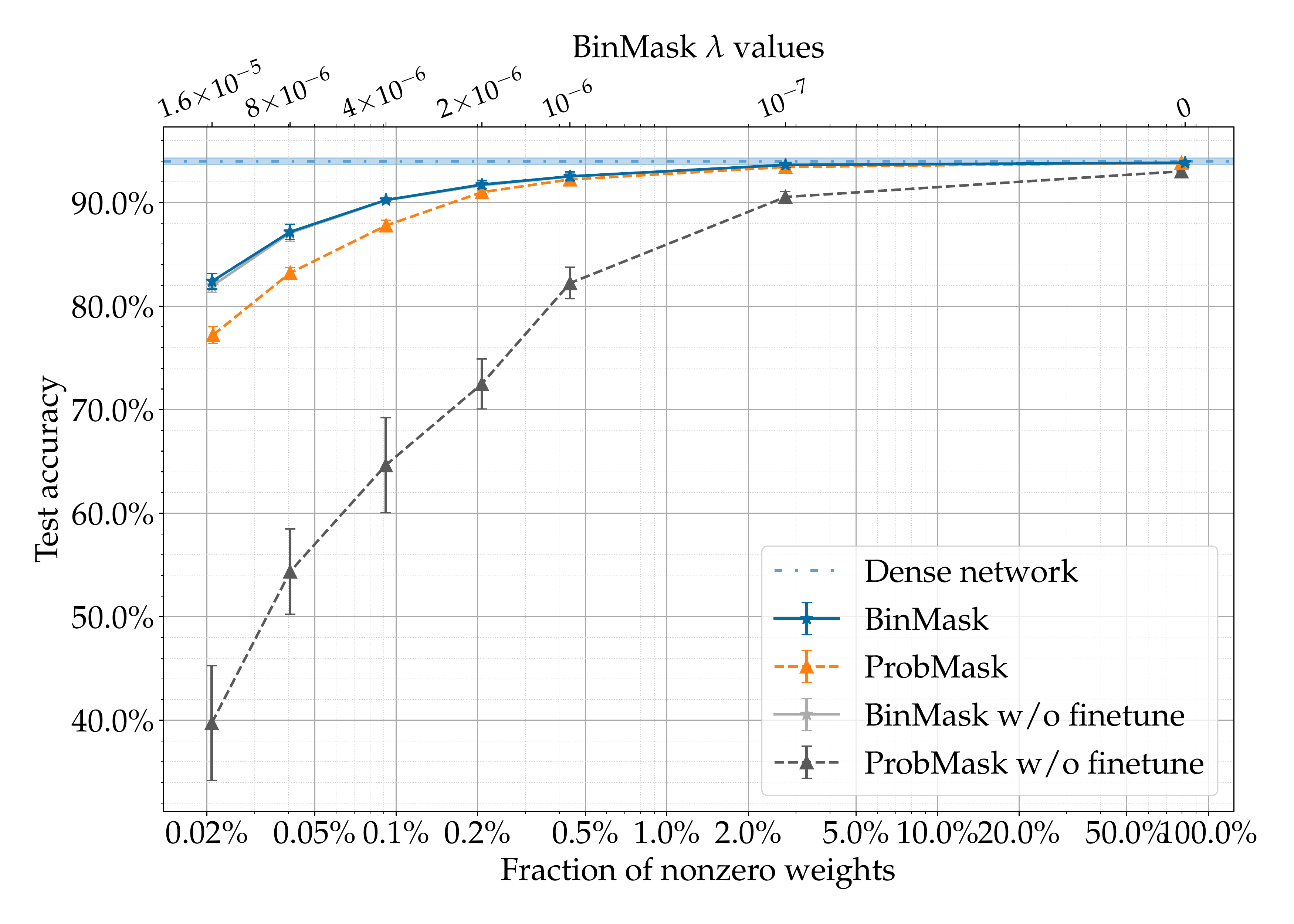}{VGG19 on CIFAR10}
    \vskip .8em
    \addSubFig{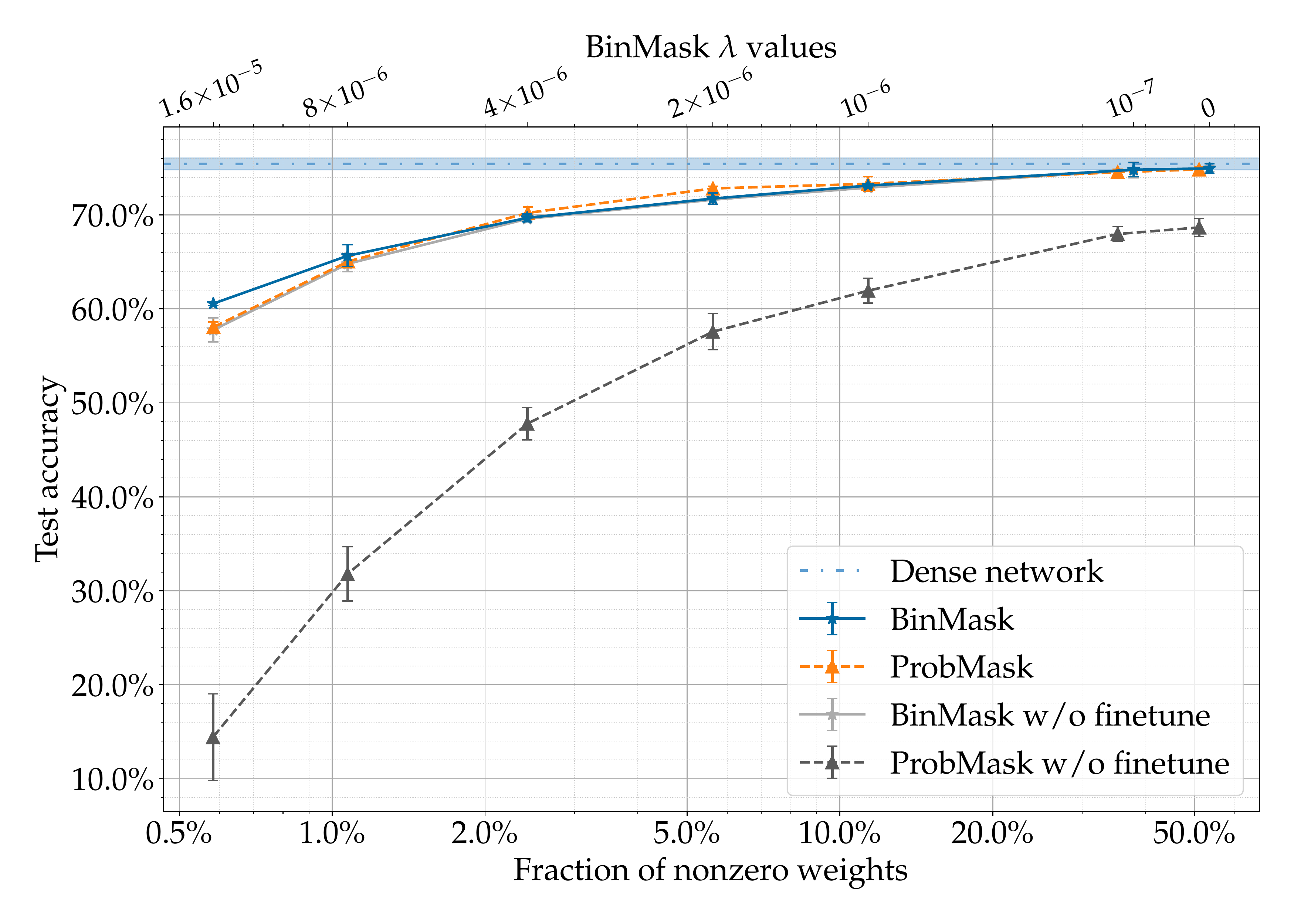}{ResNet32 on CIFAR100}
    \hfill
    \addSubFig{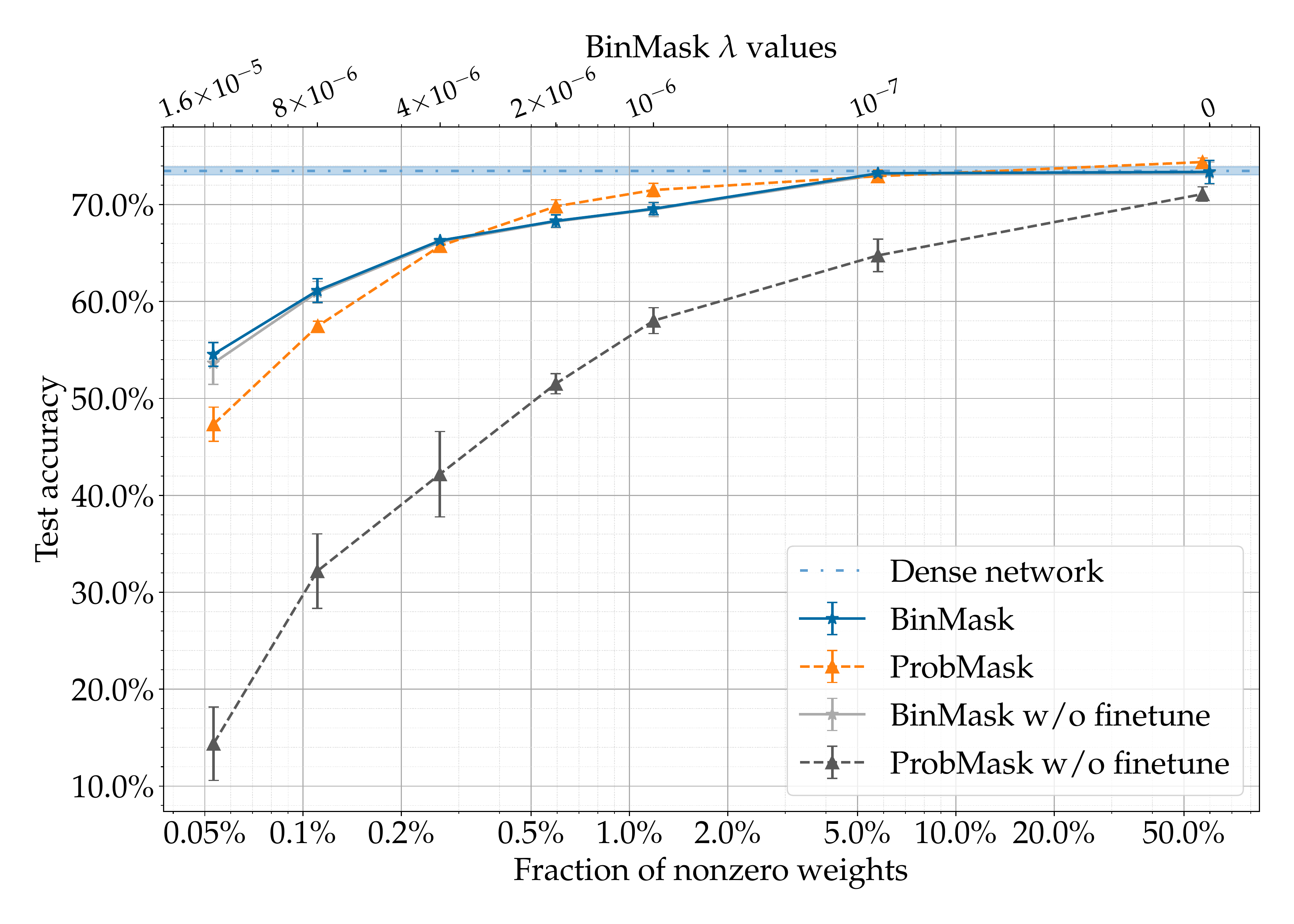}{VGG19 on CIFAR100}

    \vskip 1em
    \caption{Network sparsification results on CIFAR10/CIFAR100. Error bars and
        the transparent horizontal region indicate 95\% CI computed over four
        trials.}
    \label{fig:sparsenet}
\end{figure}

\renewcommand{\addSubFig}[2]{
    \begin{subfigure}{.31\textwidth}
        \centering
        \includegraphics[width=\textwidth]{#1}
        \vskip -.5em
        \caption{#2}
    \end{subfigure}
}
\begin{figure}[hbt]
    \centering
    \addSubFig{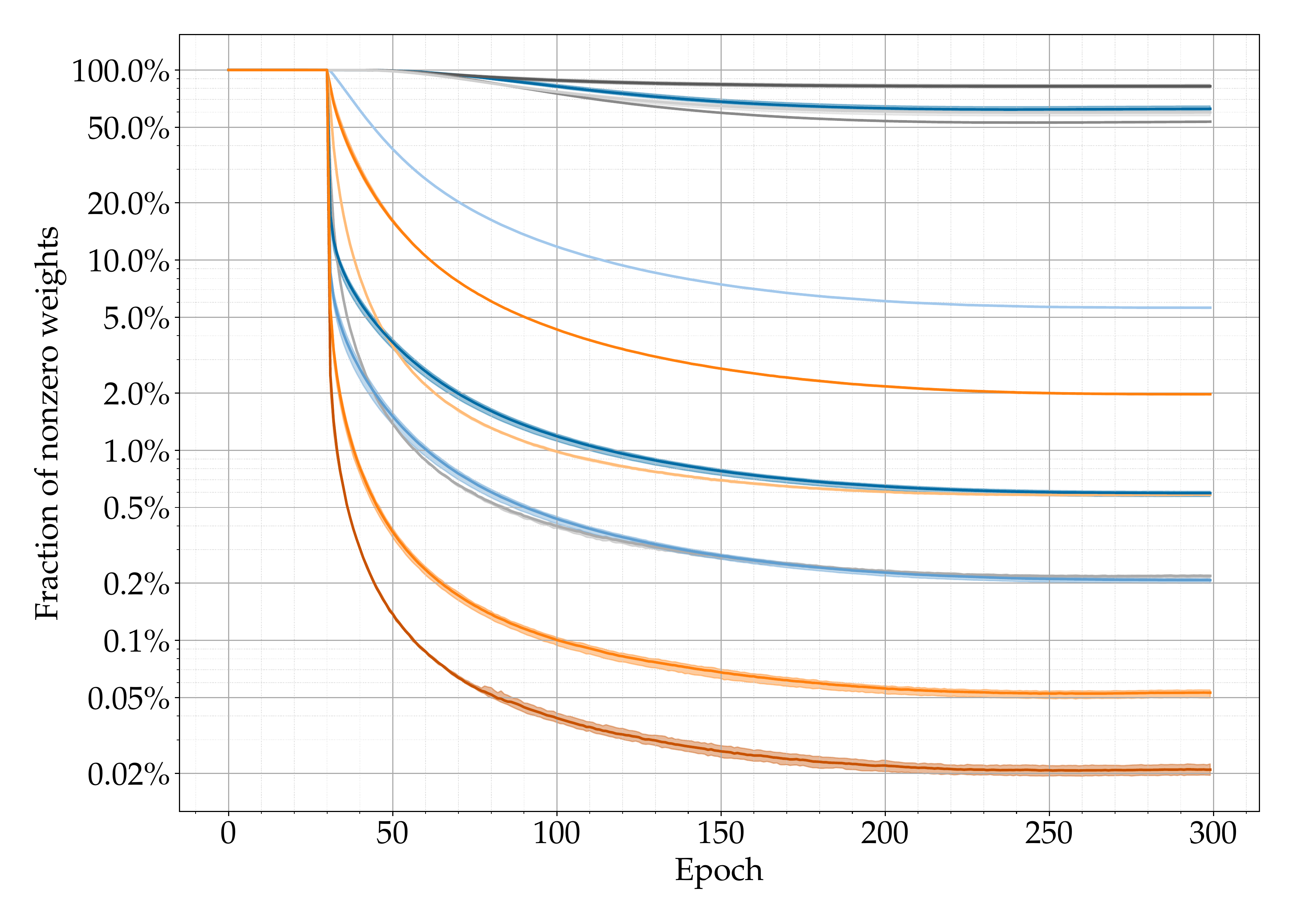}{Sparsity in log scale}
    \hfill
    \addSubFig{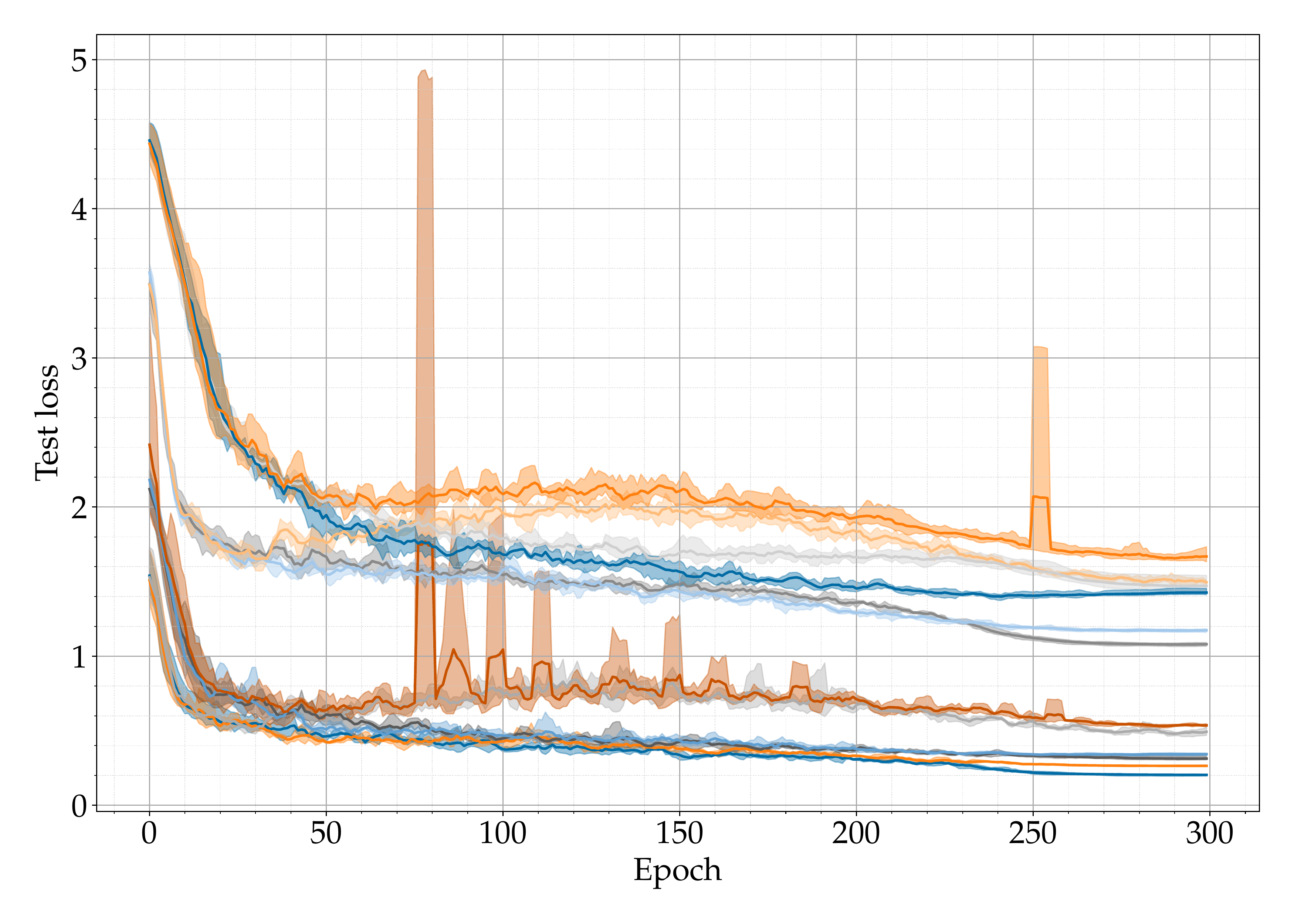}{Test loss}
    \hfill
    \addSubFig{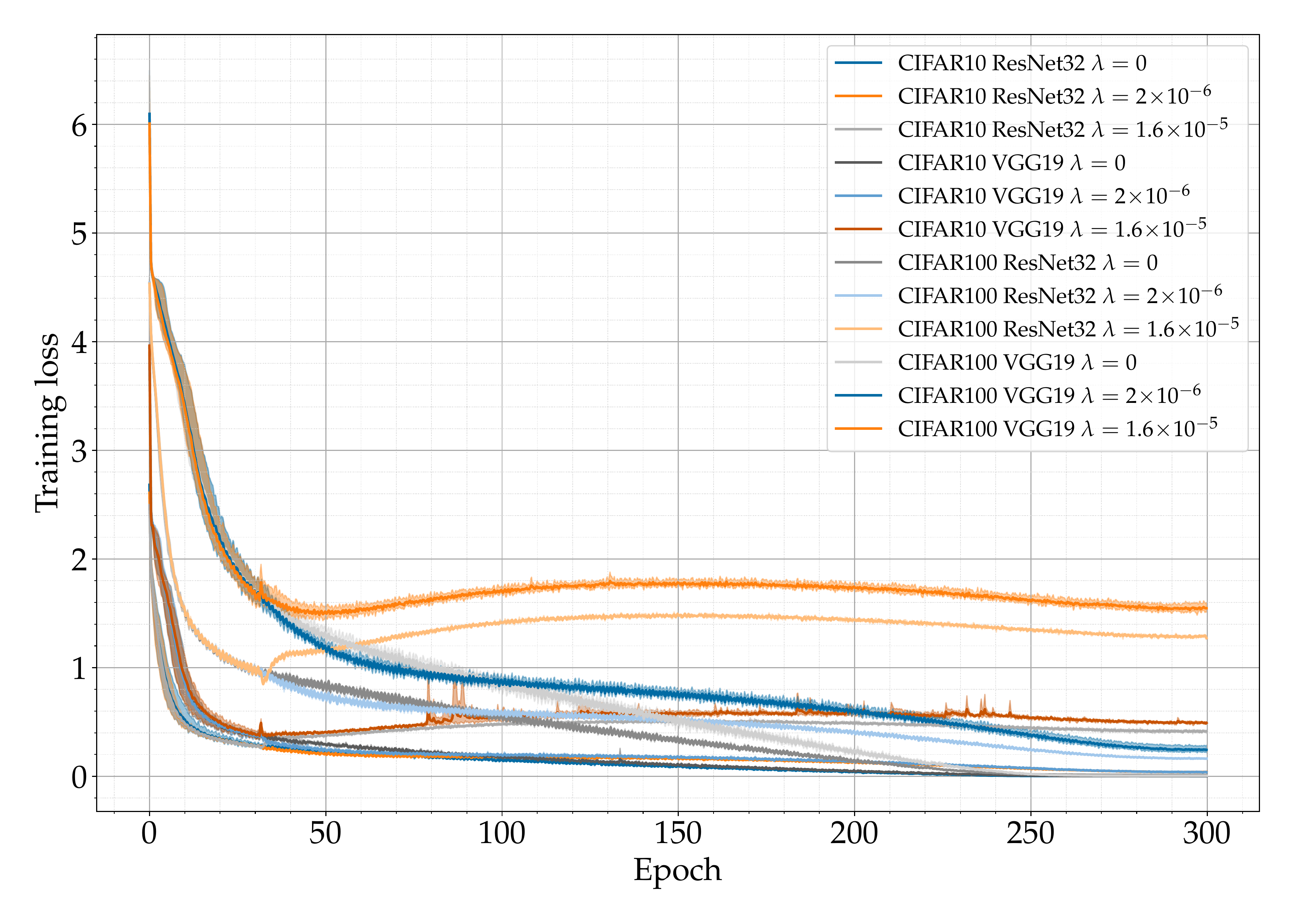}{Training loss}

    \vskip 1em
    \caption{Metrics at each training step for selected BinMask $\lambda$
        values. Filled regions of each line are min/max values over four trials.}
    \label{fig:sparsenet-dynamics}
\end{figure}

We apply BinMask to train sparse ResNet32 \citep{he2016deep} and VGG19 \citep{
simonyan2014very} networks on CIFAR-10/100 datasets \citep{
krizhevsky2009learning}. We sparsify the weights (excluding biases) of
convolutional and fully connected layers. We compare with ProbMask
\citep{zhou2021effective}, a state-of-the-art method for training sparse
networks with a probabilistic relaxation to $L_0$ constraint. We their authors'
open-source implementation%
\footnote{\url{https://github.com/x-zho14/ProbMask-official}}. ProbMask
gradually sparsifies the network during training to meet a predefined target
sparsity. We use the same setting as ProbMask for weight optimization: SGD
optimizer for 300 epochs with momentum of 0.9 and cosine learning rate annealing
starting from 0.1. The ProbMask implementation finetunes the model for 20 epochs
after training the masks. For a fair comparison, we also conduct finetuning for
the dense models and BinMask models with fixed masks and report results before
and after finetuning.

For each network architecture and dataset, we train BinMask networks with
$\lambda \in \{\sciv{1.6}{-5},\, \sciv{8}{-6},\, \sciv{4}{-6},\,
\sciv{2}{-6},\, \scivv{-6},\, \scivv{-7},\, 0\}$. We then train ProbMask
networks with the same overall sparsity as BinMask networks. For each
configuration, we run four trials and evaluate the mean accuracy on the test
set.

\cref{fig:sparsenet} presents the test accuracies. One can immediately notice
the significant performance drop experienced by ProbMask without finetuning. A
primary reason is that resampling the probabilistic masks causes shift in the
batch normalization statistics. Using per-minibatch statistics instead of the
running mean/variance in batch normalization improves ProbMask's mean test
accuracy by 12.9\% to 17.8\% in the four configurations. By contrast, BinMask
uses deterministic masks and does not exhibit a large performance drop without
finetuning. Notably, with $\lambda \in \{0,\, \scivv{-7} \}$, BinMask produces
sparse networks with up to 97\% sparsity without statistically significant
performance differences from the dense networks. BinMask delivers worse
accuracies than ProbMask in a few CIFAR-100 configurations but constantly
outperforms ProbMask in all high-sparsity cases. BinMask is computationally
efficient. On a single NVIDIA Titan XP GPU, training with BinMask is 3.40\%
slower for ResNet32 and 10.97\% slower for VGG19 than training the dense
networks. By comparison, ProbMask is 1.4x and 1.9x slower than BinMask on
ResNet32 and VGG19, respectively.

\cref{fig:sparsenet-dynamics} presents the change of sparsity, test loss, and
training loss during training. It shows that BinMask can improve sparsity
quickly during training without notably impacting training or test loss.

% f}}}

\subsection{Model regularization}
% f{{{

\begin{table}[hbt]
    \scriptsize
    \centering
    \caption{Network regularization results on the PDAC dataset.}
    \label{tab:pdac}
    \setlength{\tabcolsep}{.3em}
    \vskip 1em
    \begin{threeparttable}
    \input{gen/pdac.tex}
    \begin{tablenotes}[online]
        \item Notes:
        \begin{itemize}
            \item Columns are sorted by the test AUCs of each method.
            \item Bracketed numbers are 95\% CI computed over eight trials.
            \item LR is logistic regression.
            \item Mean weight $L_0$ is the fraction of weights whose absolute
                value is at least $\scivv{-4}$.
            \item Let $\vW\in\real^n$ be the flattened weight. For $p \ge 1$,
                mean weight $L_p$ is $\left(\frac{1}{n}\sum_{1 \le i \le n}
                \abs{\vW_i}^p \right)^{\frac{1}{p}}$.
            \item \textbf{Bold} numbers are AUCs than which the best
                training/test AUC is not significantly better ($p > 0.3$).
        \end{itemize}
    \end{tablenotes}
    \end{threeparttable}
\end{table}

The above experiments have not shown any advantage of sparse networks regarding
predictive power. This section evaluates BinMask's capability to regularize
neural networks for better generalizability. We work with a dataset for early
prediction of Pancreatic Duct Adenocarcinoma (PDAC) diagnosis, a type of
pancreatic cancer, based on Electronic Health Records (EHR) \citep{
jia2023developing}. The PDAC dataset contains features extracted from EHR of
millions of patients. The features are sparse representations of diagnosis,
medication, and lab results before a cutoff date for each patient. The outcome
variable is PDAC diagnosis within 6 to 18 months after the cutoff date. The PDAC
dataset has 63,884 positive instances and 3,604,863 negative instances, with
over five thousand features per instance (the exact number of features vary
slightly depending on training/test split because the EHR codes for feature
generation are selected based on a frequency threshold on the training set). On
average, each feature has a 94\% chance of being zero. The dataset can be
obtained under some collaboration agreement (details to be revealed after paper
review). The dataset is split into training, validation, and test sets, with
75\%, 10\%, and 15\% of the instances, respectively.

We train MLP neural networks with two hidden layers, each having 64 and 20
neurons with tanh, respectively. We use the AdamW \citep{loshchilov2017decoupled}
optimizer and cosine learning rate annealing from $0.002$ to $\sciv{5}{-5}$ with
16 epochs. We consider four regularization methods: $L_0$ regularization with
BinMask, $L_1$ regularization (a.k.a., Lasso regularization), $L_2$
regularization (a.k.a., weight decay), and dropout \citep{
srivastava2014dropout}. We test five regularization coefficients (or dropout
probabilities) for each method. For methods other than $L_2$ regularization, we
also use a weight decay of $0.01$. We adopt early stopping for all methods by
selecting the model with the highest validation AUC after each epoch.

We evaluate model performance by computing their mean AUC on the test sets over
eight trials with different dataset split and weight initialization. We also
train Logistic Regression (LR) models for reference. \cref{tab:pdac} presents
the results. Our results show that all the considered regularization methods can
effectively regularize the corresponding weight norm, but only $L_0$
regularization with BinMask and $L_1$ regularization significantly improve
network generalizability. While it has been suggested that in most clinical
cases, NN provides marginal to no improvement over LR \citep{
issitt2022classification, appelbaum2021development}, our results suggest that
with proper regularization, NN can achieve significant improvement in clinical
applications.

% f}}}

% vim: tw=80 filetype=tex foldmethod=marker foldmarker=f{{{,f}}} spell spelllang=en

%% file: gen/pdac.tex
\begin{tabular}{llrrrrr}
\toprule
Metric & Method & \multicolumn{5}{c}{Values (95\% CI)} \\
\midrule
\multirow[c]{4}{8em}{Regularization coeff. / dropout prob.} & BinMask & $3\! \times \! 10^{-5}$ & $4\! \times \! 10^{-5}$ & $2\! \times \! 10^{-5}$ & $5\! \times \! 10^{-5}$ & $10^{-5}$ \\
 & $L_1$ & $10^{-4}$ & $5\! \times \! 10^{-5}$ & $2\! \times \! 10^{-4}$ & $3\! \times \! 10^{-4}$ & $10^{-5}$ \\
 & $L_2$ & 0.01 & 0.1 & 10 & 1 & 50 \\
 & Dropout & 0.5 & 0.3 & 0.7 & 0.1 & 0.9 \\
\cmidrule(lr){1-7}
\multirow[c]{5}{*}{Training AUC} & BinMask & $0.858(\pm 0.003)$ & $0.850(\pm 0.003)$ & $0.874(\pm 0.003)$ & $0.844(\pm 0.007)$ & $0.907(\pm 0.020)$ \\
 & $L_1$ & $0.859(\pm 0.004)$ & $0.880(\pm 0.008)$ & $0.846(\pm 0.005)$ & $0.838(\pm 0.003)$ & $0.892(\pm 0.021)$ \\
 & $L_2$ & $0.879(\pm 0.028)$ & $0.890(\pm 0.038)$ & $0.831(\pm 0.006)$ & $0.914(\pm 0.041)$ & $0.796(\pm 0.004)$ \\
 & Dropout & $0.908(\pm 0.003)$ & $\bm{0.942(\pm 0.024)}$ & $0.862(\pm 0.002)$ & $0.918(\pm 0.043)$ & $0.804(\pm 0.004)$ \\
 & LR & \multicolumn{5}{c}{$0.834(\pm 0.004)$} \\
\cmidrule(lr){1-7}
\multirow[c]{5}{*}{Test AUC} & BinMask & $\bm{0.834(\pm 0.006)}$ & $0.832(\pm 0.007)$ & $0.830(\pm 0.007)$ & $0.830(\pm 0.010)$ & $0.815(\pm 0.009)$ \\
 & $L_1$ & $\bm{0.836(\pm 0.007)}$ & $0.832(\pm 0.007)$ & $0.831(\pm 0.007)$ & $0.828(\pm 0.007)$ & $0.823(\pm 0.010)$ \\
 & $L_2$ & $0.820(\pm 0.008)$ & $0.819(\pm 0.009)$ & $0.816(\pm 0.007)$ & $0.814(\pm 0.017)$ & $0.793(\pm 0.008)$ \\
 & Dropout & $0.827(\pm 0.006)$ & $0.824(\pm 0.008)$ & $0.824(\pm 0.006)$ & $0.820(\pm 0.012)$ & $0.798(\pm 0.007)$ \\
 & LR & \multicolumn{5}{c}{$0.819(\pm 0.008)$} \\
\cmidrule(lr){1-7}
\multirow[c]{4}{8em}{Mean weight $L_0$} & BinMask & $0.018(\pm 0.001)$ & $0.013(\pm 0.001)$ & $0.027(\pm 0.001)$ & $0.010(\pm 0.001)$ & $0.178(\pm 0.277)$ \\
 & $L_1$ & $0.071(\pm 0.006)$ & $0.161(\pm 0.073)$ & $0.044(\pm 0.006)$ & $0.037(\pm 0.002)$ & $0.966(\pm 0.007)$ \\
 & $L_2$ & $0.999(\pm 0.000)$ & $0.999(\pm 0.000)$ & $0.967(\pm 0.002)$ & $0.995(\pm 0.001)$ & $0.960(\pm 0.004)$ \\
 & Dropout & $0.999(\pm 0.000)$ & $0.999(\pm 0.000)$ & $0.999(\pm 0.000)$ & $0.999(\pm 0.000)$ & $0.999(\pm 0.000)$ \\
\cmidrule(lr){1-7}
\multirow[c]{4}{8em}{Mean weight $L_1$} & BinMask & $0.005(\pm 0.000)$ & $0.003(\pm 0.000)$ & $0.007(\pm 0.000)$ & $0.003(\pm 0.000)$ & $0.020(\pm 0.009)$ \\
 & $L_1$ & $0.001(\pm 0.000)$ & $0.002(\pm 0.000)$ & $0.000(\pm 0.000)$ & $0.000(\pm 0.000)$ & $0.017(\pm 0.001)$ \\
 & $L_2$ & $0.051(\pm 0.009)$ & $0.056(\pm 0.011)$ & $0.002(\pm 0.000)$ & $0.017(\pm 0.003)$ & $0.002(\pm 0.000)$ \\
 & Dropout & $0.118(\pm 0.001)$ & $0.134(\pm 0.007)$ & $0.103(\pm 0.001)$ & $0.096(\pm 0.020)$ & $0.088(\pm 0.001)$ \\
\cmidrule(lr){1-7}
\multirow[c]{4}{8em}{Mean weight $L_2$} & BinMask & $0.044(\pm 0.001)$ & $0.038(\pm 0.001)$ & $0.055(\pm 0.001)$ & $0.034(\pm 0.001)$ & $0.077(\pm 0.006)$ \\
 & $L_1$ & $0.009(\pm 0.000)$ & $0.014(\pm 0.000)$ & $0.006(\pm 0.001)$ & $0.005(\pm 0.000)$ & $0.039(\pm 0.003)$ \\
 & $L_2$ & $0.068(\pm 0.012)$ & $0.073(\pm 0.014)$ & $0.004(\pm 0.000)$ & $0.023(\pm 0.003)$ & $0.003(\pm 0.000)$ \\
 & Dropout & $0.156(\pm 0.001)$ & $0.176(\pm 0.009)$ & $0.136(\pm 0.001)$ & $0.125(\pm 0.026)$ & $0.115(\pm 0.001)$ \\
\bottomrule
\end{tabular}

%% file: conclusion.tex
\section{Conclusion}

This paper revisits a straightforward formulation, BinMask, for $L_0$
regularization of neural networks. The critical insight is decoupled
optimization of weights and binary masks. The binary mask is optimized via the
identity straight-through estimator and the Adam optimizer, which are widely
accepted choices for training binarized neural networks. We demonstrate
competitive performance of BinMask on diverse benchmarks, including feature
selection, network sparsification, and model regularization, without
task-specific hyperparameter tuning.

% vim: spell spelllang=en